\pdfoutput=1

\documentclass[11pt]{article}

\usepackage[final]{acl}
\usepackage{lineno} 

\usepackage{times}
\usepackage{latexsym}

\usepackage[T1]{fontenc}

\usepackage[utf8]{inputenc}

\usepackage{microtype}

\usepackage{inconsolata}

\usepackage{graphicx}

\usepackage{multirow}
\usepackage{algorithm}
\usepackage{algpseudocode}
\usepackage{amsmath}
\usepackage{float}  

\title{Cognitive-Aligned Document Selection for Retrieval-augmented Generation}

\author{
  Bingyu Wan\textsuperscript{1}, Fuxi Zhang\textsuperscript{1}, Zhongpeng Qi\textsuperscript{1}, Jiayi Ding\textsuperscript{1}, Jijun Li\textsuperscript{1},\\ Baoshi Fan\textsuperscript{1}, Yijia Zhang\textsuperscript{1}, and Jun Zhang\textsuperscript{1} \\ \\
  \textsuperscript{1}Dalian Maritime University \\
  \texttt{forrestwby@dlmu.edu.cn}
}

\begin{document}
\nolinenumbers

\maketitle
\begin{abstract}
Large language models (LLMs) inherently display hallucinations since the precision of generated texts cannot be guaranteed purely by the parametric knowledge they include. Although retrieval-augmented generation (RAG) systems enhance the accuracy and reliability of generative models by incorporating external documents, these retrieved documents often fail to adequately support the model's responses in practical applications. To address this issue, we propose GGatrieval (Fine-\textbf{G}rained \textbf{G}rounded \textbf{A}lignment Re\textbf{trieval} for verifiable generation), which leverages an LLM to dynamically update queries and filter high-quality, reliable retrieval documents. Specifically, we parse the user query into its syntactic components and perform fine-grained grounded alignment with the retrieved documents. For query components that cannot be individually aligned, we propose a dynamic semantic compensation mechanism that iteratively refines and rewrites the query while continuously updating the retrieval results. This iterative process continues until the retrieved documents sufficiently support the query's response. Our approach introduces a novel criterion for filtering retrieved documents, closely emulating human strategies for acquiring targeted information. This ensures that the retrieved content effectively supports and verifies the generated outputs. On the ALCE benchmark, our method significantly surpasses a wide range of baselines, achieving state-of-the-art performance.
\end{abstract}

\section{Introduction}

Retrieval-augmented generation (RAG) combines retrieval mechanisms with large language models (LLMs) to address the limitations of purely generative models \citep{ref6}. By retrieving external information relevant to user queries through retrieval mechanisms, RAG enhances generated outputs \citep{ref60,min-etal-2020-ambigqa} and mitigates the hallucination problem in LLMs \citep{ref71}, allowing outputs to incorporate up-to-date real-world information \citep{ref69}, often without requiring additional training \citep{ref70}.

\begin{figure} 
  \centering 
  \includegraphics[width=\columnwidth]{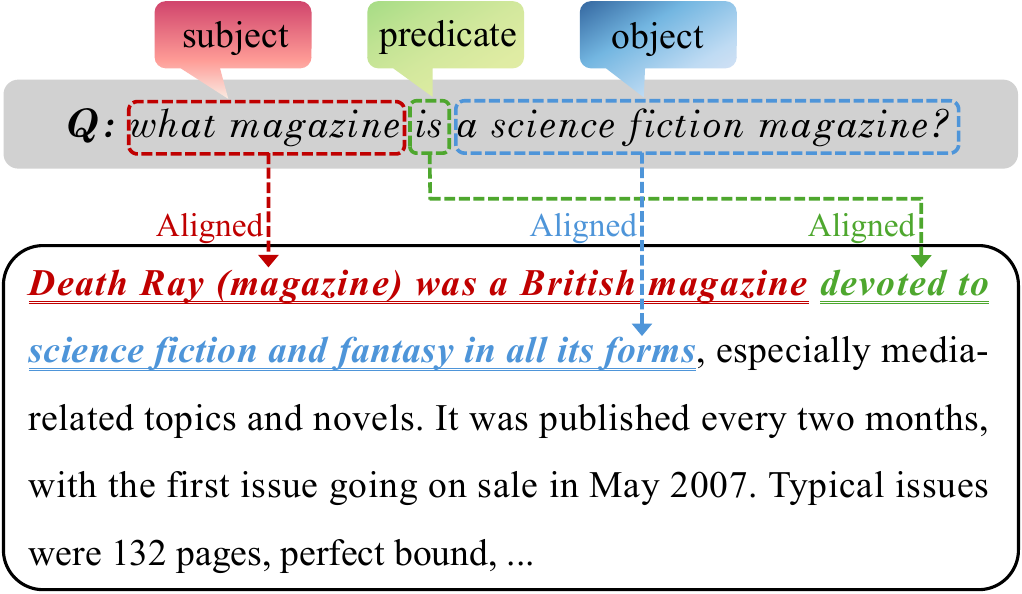} 
  \caption{An example for our criterion. The example shows that for each syntactic component of the query, a corresponding grounded segment in the retrieved document can be identified that aligns with it.}
  \label{fig:paper1} 
\end{figure}

Despite these advancements, the performance of current RAG systems remains constrained by several factors, such as the quality of retrieved documents, prompt design, and the scale of the generative model. Among these, the quality of retrieved documents is pivotal in determining system performance \citep{ref55}. High-quality retrieval is mainly dependent on the effectiveness of the retrieval mechanism. Previous research has proposed various optimization strategies to enhance retrieval mechanisms, including recursive retrieval for richer content \citep{li-etal-2024-llatrieval,ref72}, adjusting corpus chunk sizes for optimized outcomes \citep{ref73,raina-gales-2024-question,wang-etal-2024-rag}, fine-tuning retrievers \citep{ref31,zhang-etal-2024-arl2,ref32,ref70,ref71}, re-ranking retrieved content for diversity and quality \citep{saad-falcon-etal-2023-udapdr,ref76,sun-etal-2023-chatgpt}, and rephrasing retrieved text to better leverage generative models \citep{ref66,ref77}. Beyond retrieval strategies,  verifiable generation \citep{gao-etal-2023-enabling,li-etal-2024-llatrieval} has emerged as a novel paradigm that enables LLMs to generate cited text while assessing the quality of retrieved information. This paradigm provides a valuable reference for reproducing and comparing different retrieval mechanisms.  

However, most existing research focuses on refining retrieval mechanisms through advanced  technologies, such as clustering, retrievers, and LLMs, without addressing the need for effective selection criteria to select supportive documents. The absence of such criteria often leads to additional noisy documents, negatively affecting the reliability and verifiability of generated outputs.

To address this challenge, we propose \textbf{GGatrieval} (Fine-\textbf{g}rained \textbf{G}rounded \textbf{A}lignment Re\textbf{trieval} for Verifiable Generation), a framework that optimizes retrieval mechanisms by leveraging the verifiable generation paradigm. As part of this framework, we introduce a novel criterion for evaluating document quality. This criterion draws inspiration from the cognitive process of human document selection, where a retrieved document must contain a continuous segment that semantically aligns with each syntactic component of the query, as illustrated in Figure 1. This criterion forms the foundation for our document classification approach and guides the optimization of two key stages in the retrieval process: (1) Retrieval stage: To obtain more reliable retrieved documents, we design a Semantic Compensation Query Update (\textbf{SCQU}) strategy, inspired by GenRewrite \citep{ref42} and Rewrite-Retrieve-Read \citep{ma-etal-2023-query}. This strategy initially formulates synonymous queries to address gaps in the document relative to the query. Subsequently, it generates pseudo-documents tailored to these synonymous queries, ensuring alignment with the proposed criterion. Finally, the method concatenates the synonymous queries and their corresponding pseudo-documents to construct an updated query for retrieving additional candidate documents. This strategy effectively leverages the retriever's ability to retrieve semantically similar information, enhancing retrieval reliability by generating queries semantically aligned with the target documents. (2) Re-ranking Stage: To obtain more verifiable retrieved documents, we propose a Fine-grained Grounded Alignment (\textbf{FGA}) strategy in which a LLM performs syntactic parsing on the query, extracting its grammatical components. It then aligns these components with the document to generate grounded alignment labels, which are crucial for document re-ranking. By improving the verifiability of the retrieved supporting documents, this approach further improves the reliability of RAG systems.

We evaluate the quality of retrieved documents and the overall system performance using the ALCE benchmark \citep{gao-etal-2023-enabling} and an extended Natural Questions dataset \citep{kwiatkowski-etal-2019-natural}. Our experiments demonstrate that GGatrieval outperforms mainstream baselines, setting new performance standards. Notably, on the ELI5 dataset, our method improves the Claim F1 and Citation F1 scores by 22\% and 28\%, respectively.  

The main contributions of this paper are as follows:

\noindent(1) To the best of our knowledge, we are the first to introduce a document selection criterion based on human judgment, enabling effective categorization and filtering of retrieved documents. 

\noindent(2) We introduce the SCQU strategy to bridge semantic gaps between queries and target documents, improving retrieval quality and efficiency. 

\noindent(3) We present the FQA strategy, enhancing verifiable document retrieval in RAG systems. 

\noindent(4) We conduct experiments on four diverse datasets and across representative baselines from distinct retrieval optimization mechanisms, achieving state-of-the-art performance.  

\section{Related Work}
\subsection{Verifiable Generation}

\begin{figure*}[t]
  \centering 
  \includegraphics[width=\linewidth]{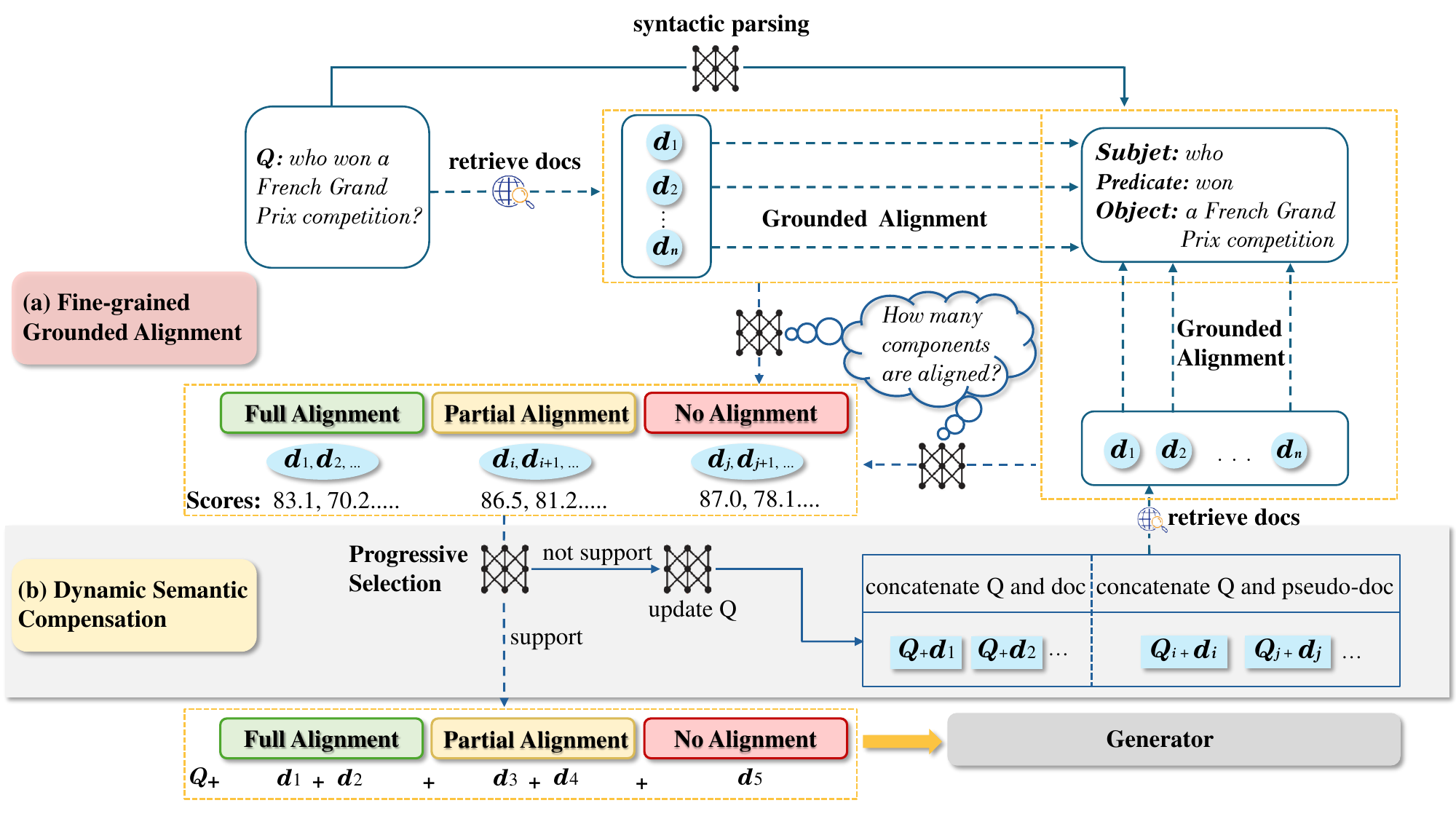} 
  \caption{Overview of our GGatrieval. Our approach (Section \ref{Section3.4}) leverages large language models for the Fine-grained Grounded Alignment strategy (Section \ref{Section3.2}) to obtain document labels {Full Alignment, Partial Alignment, and No Alignment} (Section \ref{Section3.1}), while implementing the Dynamic Semantic Compensation strategy (Section \ref{Section3.3}) for query updates to enhance the retrieval of highly aligned documents.}
  \label{fig:figure_2} 
\end{figure*}

Verifiable generation refers to the process of generating text that can be independently verified, ensuring users can trace the source of each piece of knowledge. Existing paradigms of verifiable generation can be broadly categorized into two approaches: (1) Direct generation of responses with citations: These methods aim to enhance content verifiability by augmenting the generative capabilities of language models, allowing them to incorporate citation information directly during generation. For example, \citet{weller-etal-2024-according} utilizes the large language model's intrinsic abilities to generate citations by prompting it with phrases like "according to Wikipedia". \citet{ref22} critically assesses the quality of generated text and provided feedback based on these evaluations, guiding the language model towards improvement. (2) Retrieval-based responses with citations: This approach enhances citation accuracy by retrieving external information, such as web pages or documents, to supplement generated content. Notable examples include WebGPT \citep{ref26} and LaMDA \citep{ref28}, which use web and Wikipedia data to construct large-scale pretraining datasets, enabling citation-inclusive responses. Moreover, \citet{li-etal-2024-llatrieval} iteratively improves the quality of citations in the generated text by continually updating the retrieval results until the retrieved information aligns with the generated answers. Our approach primarily adopts the second paradigm, facilitating a more effective comparison of various retrieval mechanisms within RAG systems.

\subsection{Optimizing retrieval mechanisms for RAG}
In retrieval-augmented question-answering systems, the reliability of the retrieved documents is crucial for ensuring both the accuracy and trustworthiness of the generated responses. The technologies for optimizing retrieval mechanisms can be broadly classified into two categories: (1) Model-training Retrieval Augmentation: CRAG \citep{ref31} and DR-RAG \citep{ref32} enhance the quality of documents by training retrievers and classifiers to evaluate and filter the retrieved documents. Additionally, \citet{ref33} develops a query rewriting model that incorporates feedback from a re-ranker. Self-Retrieval \citep{ref34} allows large models to conduct self-retrieval on pre-trained datasets by incorporating retrieval corpora into their training. (2) LLM-based Retrieval Augmentation: In query rewriting, \citet{ref41} introduces Context-Aware Query Rewriting, which leverages LLMs for query understanding in text ranking tasks, thereby improving the accuracy and relevance of the query. For document re-ranking, Neural PG-RANK \citep{ref43} optimizes decision quality metrics using the Plackett-Luce ranking strategy, incorporating LLMs into the training process. Similarly, RankGPT \citep{sun-etal-2023-chatgpt} uses ChatGPT-generated document ranking to reorder documents effectively. Furthermore, in terms of pipeline optimization, LLatrieval \citep{li-etal-2024-llatrieval} combines query rewriting and re-ranking through an iterative updating mechanism, continuously refining the retrieval results and verifying that the documents retrieved are sufficient to support the generated answers until the validation criteria are met.

In contrast to recent related studies \citep{ref31,sun-etal-2023-chatgpt,li-etal-2024-llatrieval}, our work focuses on the alignment criterion between retrieved documents and user queries, addressing the reliability of retrieved documents from the perspective of human cognition, thereby further enhancing system performance.

\section{Methods}

Figure 2 shows the overview of GGatrieval at inference.  
\subsection{Document Toxonomy}\label{Section3.1}
The meaning of complex expressions is constructed from the meanings of their simpler components \citep{ref1}. Analyzing syntactic structure is critical in understanding sentence meaning \citep{lesmo-lombardo-1992-assignment}. Therefore, the cognitive process of human document retrieval can be conceptualized as follows:  (1) decomposing the query into its grammatical components; (2) identifying the corresponding segments within the retrieved documents that align semantically with the query's components; and (3) classifying the document as high-quality if all components of the query align with the retrieved text. Based on this cognitive process, we propose a selection criterion for evaluating retrieval documents: whether there are continuous grounded segments within the document that align with the query's grammatical components. This criterion categorizes retrieval documents into three types:

\noindent \textbf{Full Alignment:}  A continuous segment in the document aligns semantically with all query components.

\noindent \textbf{Partial Alignment:} A continuous segment in the document aligns semantically with at least one component of the query but not all.

\noindent \textbf{No Alignment:} No continuous segment in the document semantically aligns with any query component.

The retrieval mechanism of GGatrieval prioritizes documents labeled with Full Alignment, while those labeled with Partial Alignment contribute to system robustness. Documents with No Alignment serve as exclusion criteria in the retrieval process. 

\subsection{Fine-grained Grounded Alignment}\label{Section3.2}
In GGatrieval, filtering reliable and verifiable documents requires the acquisition of alignment labels for the retrieved documents. To achieve this, we design the FGA strategy. Given the strong performance of LLMs in tasks like syntactic analysis and natural language understanding \citep{tian-etal-2024-large,zhou-etal-2024-grasping,ref3}, we utilize LLMs to implement this strategy. The procedure is outlined as follows:

\noindent \textbf{Query Syntactic Analysis:} The system processes the user's query \( Q \) and decomposes it into key syntactic components, including the subject, predicate, and object, among others, as described in Equation (1): 
\begin{equation}
\begin{alignedat}{2}
C &= \left\{C_{s},C_{v},C_{o},C_{c},C_{attr},C_{adv},C_{supp},C_{app} \right\} \\
  &= f_{\text{LLM}}(Q)
\end{alignedat}
\end{equation} 
\noindent \sloppy
Here, \( C \)  represents the set of syntactic components extracted from the query \( Q \),
where \(C_{s}\), \(C_{v}\), \(C_{o}\), \(C_{c}\), \(C_{attr}\), \(C_{adv}\), \(C_{supp}\) and \(C_{app}\) represent the subject, predicate, object, predicative, attributive, adverbial, complement, and apposition, respectively.

\noindent \textbf{Fine-grained Grounded Alignment:} The system compares each syntactic component of the query with the candidate document \(D\), producing a list of query components that can be semantically aligned with the document. This alignment is mathematically represented as: 
\begin{equation} 
M(Q, D) = \{ C_i \mid C_i \in C, \, f_{\text{LLM}}(C_i, D) = 1 \} 
\end{equation}
In this expression, \(f_{LLM}(C_i, D)\) denotes the alignment function computed by the LLM, which outputs 1 if alignment is found or 0 if no alignment is detected. The set \(M(Q,D)\) contains the query components that align semantically with the document. The greater the number of aligned components, the higher the probability that the candidate document answers the user's query, making this list vital for filtering candidate documents.

\noindent \textbf{Reflection and Query Optimization:} As highlighted by \citet{ref4}, the self-reflection abilities of LLMs can enhance their performance. Accordingly, we incorporate a reflection step where the LLM generates a reflective list of aligned components \(M'(Q,D)\). For any unaligned components \(C_i \in C\), the LLM generates synonymous descriptions \(C'_i\) and updates the query \(Q'\) as follows:
\begin{equation}
Q' = \left\{ 
\begin{split}
C_i' = \text{Synonym}(C_i) \\
\text{if} \ f_{\text{LLM}}(C_i,P) = 0 
& \quad \text{else} \quad C_i 
\end{split}
\right\}
\end{equation}
where \(C'_i\) represents the synonymous description of the component \(C_i\). The updated query \(Q'\), now incorporating these synonymous descriptions, forms the foundation for the subsequent phase of the SCQU (Section \ref{Section3.3}).

\begin{algorithm}
\caption{GGatrieval}
\label{alg:document_retrieval}
\small  
\textbf{Input:} Question $q$, document pool $D_c$, reranked document pool $D_o$, the large language model LLM, the Retriever $R$, the maximum iteration 
$T$, each iteration's document candidates quantity $N$ \\
\textbf{Output:} Supporting Documents $D$ 
\begin{algorithmic}[1] 
\State $Q \gets q$
\State $D \gets \{ \}$
\State $C \gets \{C_{s},C_{v},C_{o},C_{c},C_{attr},C_{adv},C_{supp},C_{app}\}$
\State $ \quad\quad\,\,= f_{LLM}(Q)$ 

\For{$i \in (1, T)$}
  \If{$D \neq \left\{ \right\}$}
      \State $Q \gets$ SCQU strategy
  \EndIf

  \State $D_c \gets R(Q, N) $
    
  \For{$D_c^* \in D_c$}
    \State $D_c^* \gets$ FGA strategy
  \EndFor
    \State $D_o \gets$ Rerank $D_c$ with alignment label
  \For{$D_o^* \in$ SlidingWindow($D_o$)}
    \State $D \gets$ Use the LLM to select $k$ docs from $D \cup D_{o}^{*}$
  \EndFor

  \If{Verify($q, D$) $\rightarrow$ Yes}
    \State \textbf{break}
  \EndIf
\EndFor

\State \textbf{Return} $D$
\end{algorithmic}
\end{algorithm}

\subsection{Semantic Compensation Query Update}\label{Section3.3}

Dense retrieval models effectively retrieve documents semantically related to a given query \citep{karpukhin-etal-2020-dense,ref6}. Building on this property, we design the SCQU strategy as follows:

\noindent \textbf{Categorizing candidate documents:} Candidate documents are classified based on \(r\), the proportion of syntactic components in the query aligned by the document. Documents are categorized into two types: high-alignment documents (where \(r >= \tau\)) and low-alignment documents (where \(r<\tau\)), with \(\tau\) being a user-defined threshold.

\noindent \textbf{Updating query:} For high-alignment documents D, the system concatenates the current query \(Q\) with the document \(D\) to generate an updated query \(Q''\). This operation is based on the premise that documents with high alignment are likely to yield even better results when the semantic content of the query is enriched.
For low-alignment documents, the system generates a pseudo-document \(D_{pseudo}\) semantically aligned with the suggested query \(Q'\) (Section \ref{Section3.2}). This pseudo-document is then concatenated with \(Q'\), thereby enriching the query with relevant semantic information. As a result, the updated query facilitates the retrieval of documents that are more closely aligned with the original query. The final query update strategy is expressed as follows:
\begin{equation}
Q''=\left\{\begin{array}{ll}
Q+D, & \text { if } \frac{|M'(Q, D)|}{ |C|} \geq \tau \\
Q'+D_{\text {pseudo }}, & \text { if } \frac{|M'(Q, D)|}{|C|} < \tau
\end{array}\right.
\end{equation}
This approach dynamically updates the query \(Q\) in each retrieval cycle based on alignment conditions. Compared to LLatrieval \citep{li-etal-2024-llatrieval}, Our method reduces document retrieval by 95\% on the ASQA dataset and 67\% on the QAMPARI dataset, significantly enhancing retrieval efficiency. The number of documents retrieved by the two methods is provided in Appendix B.

\subsection{Align-update Iteration}\label{Section3.4}
Algorithm 1 outlines the workflow of our system. The process begins with parsing the user query into syntactic components \(C=\{C_{s},C_{v},C_{o},C_{c},C_{attr},C_{adv},C_{supp},C_{app}\}\) (Line 3\(\sim\)4), initiating the iterative process. In each iteration, the system applies the SCQU strategy to retrieve a refined set of candidate documents \(D_c\) (Lines 6\(\sim\)9), ensuring that the retrieved documents are increasingly semantically aligned with the query. Next, the FGA strategy is employed to assign alignment labels to each document in \(D_c\) (Lines 10\(\sim\)11). These documents are then reordered based on the alignment labels and relevance, resulting in a prioritized set \(D_o\) (Line 13), which includes documents that meet the verifiability criteria. Finally, the system employs the Progressive Selection and Document Verification methods proposed by \citet{li-etal-2024-llatrieval} to select and validate the final supporting documents (Lines 14\(\sim\)18). GGatrieval defines a robust selection criterion to establish clear retrieval objectives. Through the align-update iteration, the retrieval results are progressively refined to yield documents that better align with the retrieval goal, thereby enabling the LLM to generate both accurate and verifiable answers.

\begin{table*}
  \centering
  \resizebox{\linewidth}{!}{ 
  \begin{tabular}{llcccccccccccc}
    \hline
    \multicolumn{2}{c}{Dataset} & \multicolumn{4}{c}{ASQA} & \multicolumn{4}{c}{QAMPARI} & \multicolumn{4}{c}{ELI5} \\ \hline
    & \multicolumn{1}{c}{} & Correct & \multicolumn{3}{c}{Citation} & Correct & \multicolumn{3}{c}{Citation} & Correct & \multicolumn{3}{c}{Citation} \\ \cline{3-14} 
    & \multicolumn{1}{c}{} & EM-R & Rec & Prec & F1 & F1 & Rec & Prec & F1 & Claim & Rec & Prec & F1 \\ \cline{3-14} 
    & \multicolumn{1}{c}{} & \multicolumn{12}{c}{Meta-Llama3-8B-Instruct} \\ \hline
    \multirow{2}{*}{Vanilla} & BM25 & 31.03 & 23.5 & 24.58 & 24.02 & 5.97 & 7.34 & 8.25 & 7.76 & 10.88 & 23.93 & 26.56 & 25.18 \\
    & BGE-E-large & 40.59 & 33.83 & 37.79 & 35.7 & 6.55 & 11.62 & 13.88 & 12.62 & - & - & - & - \\
    Model training & CRAG & 33.92 & 33.86 & 36.28 & 35.05 & 5.91 & 7.0 & 8.34 & 7.6 & 11.19 & 29.19 & 32.16 & 30.61 \\
    \multirow{2}{*}{LLM-based} & RankGPT & 40.17 & 36.59 & 38.68 & 37.61 & 9.24 & 16.8 & 19.7 & 18.13 & 11.2 & 24.89 & 28.52 & 26.58 \\
    & LLatrieval & 40.58 & \textbf{40.13} & \textbf{42.94} & \textbf{41.5} & 6.93 & 13.7 & 14.52 & 14.09 & 11.23 & 27.96 & 32.9 & 30.2 \\
    Ours & GGatrieval & \textbf{41.37} & 39.7 & 42.92 & 41.22 & \textbf{9.5} & \textbf{18.06} & \textbf{21.13} & \textbf{19.47} & \textbf{11.64} & \textbf{30.41} & \textbf{33.7} & \textbf{32.0} \\ \hline
    &  & \multicolumn{12}{c}{gpt-3.5-turbo} \\ \hline
    \multirow{2}{*}{Vanilla} & BM25 & 36.57 & 36.79 & 38.98 & 37.86 & 10.21 & 18.95 & 19.67 & 19.3 & 11.65 & 41.28 & 41.4 & 23.47 \\
    & BGE-E-large & 51.57 & 53.63 & 55.95 & 54.73 & 12.06 & 26.42 & 27.48 & 26.93 & - & - & - & - \\
    Model training & CRAG & 46.29 & 47.15 & 50.19 & 48.59 & 11.9 & 20.1 & 20.73 & 20.41 & 12.79 & 48.07 & 48.96 & 48.49 \\
    \multirow{2}{*}{LLM-based} & RankGPT & 49.76 & 51.48 & 54.71 & 53.04 & 16.4 & 33.1 & 34.24 & 33.66 & 11.6 & 42.38 & 43.12 & 42.74 \\
    & LLatrieval & 50.8 & 53.54 & 55.75 & 54.58 & 16.86 & 34.09 & 34.9 & 34.46 & 11.62 & 43.47 & 44.85 & 43.88 \\
    Ours & GGatrieval & \textbf{52.86} & \textbf{56.93} & \textbf{58.19} & \textbf{57.51} & \textbf{17.85} & \textbf{35.44} & \textbf{36.58} & \textbf{35.98} & \textbf{14.21} & \textbf{56.16} & \textbf{56.26} & \textbf{56.17} \\ \hline
  \end{tabular}
  }
  \caption{The performance comparison on ALCE \citep{gao-etal-2023-enabling}. The results are categorized according to different types of retrieval technologies, with the bolded numbers indicating the best performance among all methods.}
  \label{tab1}
\end{table*}

\section{Experiment}
\subsection{Experimental Setting}
\noindent \textbf{Datasets.} We evaluate our approach using the ALCE benchmark \citep{gao-etal-2023-enabling}. ASQA \citep{stelmakh-etal-2022-asqa} is a long-form factoid question answering (QA) dataset in which each query requires multiple short answers to address its multifaceted nature. QAMPARI \citep{amouyal-etal-2023-qampari} is also a fact-based QA dataset where answers consist of entity lists extracted from multiple documents. ELI5 \citep{fan-etal-2019-eli5} is a long-form QA dataset requiring detailed, comprehensive answers with evidence from multiple documents. We also include the Natural Questions (NQ) dataset \citep{kwiatkowski-etal-2019-natural} to provide a more comprehensive assessment. NQ is a widely used benchmark for open-domain QA proposed by Google Research.

\noindent \textbf{Baselines.} We first compare it with two representative retrieval methods. Given that our focus is optimizing retrieval mechanisms within the verifiable generation framework, we select three representative methods from two primary technologies to highlight GGatrieval's advantage in enhancing retrieval efficiency and accuracy. Specifically, the Vanilla Retrievers include BM25 \citep{ref17} and BGE-large \citep{ref52}, while the model-training retrieval augmentation method includes CRAG \citep{ref31}. We use the CRAG model to filter all documents retrieved using the GGatrieval approach to ensure a fair comparison. Additionally, we consider LLM-based retrieval augmentation methods, represented by RankGPT \citep{sun-etal-2023-chatgpt} and LLatrieval \citep{li-etal-2024-llatrieval}.

\noindent \textbf{Evaluation Metrics.} We evaluate GGatrieval's performance across the four datasets following the evaluation guidelines from ALCE \citep{gao-etal-2023-enabling}. (1) For ASQA, we use Exact Match Recall (EM-R) to assess the correctness of the answers. (2) For QAMPARI and NQ, we calculate the F1 score by identifying exact matches to the gold answer list. (3) For ELI5, we measure whether the model’s predictions entail the sub-claims in the gold answers, as described in \citet{gao-etal-2023-enabling}. Additionally, we assess the verifiability of the answers by measuring citation quality, for more details about citation quality, please refer to Appendix A.

\noindent \textbf{Implementation Details.} We utilize the verifiable generation \citep{gao-etal-2023-enabling,li-etal-2024-llatrieval} paradigm for both answer generation and evaluation, aiming to assess the verifiability and accuracy of the generated responses and compare the effectiveness of various retrieval mechanisms. The APIs of OpenAI's "gpt-3.5-turbo" language model and the open-source "Meta-Llama3-8B-Instruct" model are used for grounded alignment, query updates, and generation, with the temperature set to 0 to minimize random variation. The number of supporting documents is set to 5, and the maximum number of iterations is 4. For ASQA, QAMPARI, and NQ datasets, the retrieval corpus is based on the Wikipedia dataset used in ALCE \citep{gao-etal-2023-enabling}, with the dense embedding model BGE-large \citep{ref15} as the retriever. For the ELI5 dataset, we use the Sphere \citep{ref16} corpus and followed ALCE \citep{gao-etal-2023-enabling}, employing BM25 \citep{ref17} for document retrieval due to the higher cost and slower speed of dense retrievers on large-scale web corpora. For more details about the experimental setting, please refer to Appendix A.

\subsection{Main Results}
Table 1 presents experimental results across three datasets, using two LLMs—"gpt-3.5-turbo" and "Meta-Llama3-8B-Instruct"—as base models. The choice of these two distinct models, along with representative baselines from distinct technologies, is designed to assess the generalization ability and effectiveness of GGatrieval across diverse scenarios.  

\noindent \textbf{GGatrieval consistently outperforms baseline methods on all three datasets.} Specifically, the substantial improvement in Citation F1 underscores GGatrieval's ability to retrieve high-quality documents, while the increase in Correctness highlights its positive impact on the overall performance of the RAG system. Notably, on the ELI5 dataset, GGatrieval achieves improvements of 22\% and 28\% in these two metrics, respectively. A key strength of GGatrieval over the baseline methods is its ability to evaluate and filter documents based on specific criteria, validating both the necessity and effectiveness of robust selection criteria. 

\noindent \textbf{GGatrieval consistently outperforms baseline methods across distinct base models.} Further analysis reveals that the "gpt-3.5-turbo" model consistently outperforms the "Meta-Llama3-8B-Instruct" model across all methods, confirming the scaling laws principle \citep{ref53}. Moreover, GGatrieval performs best on both base models, demonstrating its superior generalization ability. This capability can be attributed to the optimized retrieval mechanism of GGatrieval, which leverages the characteristics of the retriever to design query update strategies and simulates the human thought process in selecting retrieved documents, thereby enhancing overall system performance.

\begin{table}[H]
  \centering
  \resizebox{\linewidth}{!}{ 
  \begin{tabular}{llcccc}
    \hline
    \multicolumn{2}{c}{Dataset} & \multicolumn{4}{c}{NQ} \\ \hline
    & \multicolumn{1}{c}{} & Correct & \multicolumn{3}{c}{Citation} \\ \cline{3-6} 
    & \multicolumn{1}{c}{} & F1 & Rec & Prec & F1 \\ \hline
    \multirow{2}{*}{Vanilla} & BM25 & 20.08 & 31.28 & 33.61 & 32.41 \\
    & BGE-E-large & 26.06 & 31.5 & 34.95 & 33.13 \\
    Model training & CRAG & 23.98 & 23.7 & 26.19 & 24.88 \\
    \multirow{2}{*}{LLM-based} & RankGPT & 27.83 & 33.45 & 36.76 & 35.03 \\
    & LLatrieval & 27.3 & 32.64 & 36.03 & 34.25 \\
    Ours & GGatrieval & \textbf{28.68} & \textbf{34.51} & \textbf{37.58} & \textbf{35.98} \\ \hline
  \end{tabular}
  }
  \caption{Performance comparison on NQ with gpt-3.5-turbo. The bolded numbers indicate the best performance.}
  \label{tab2}
\end{table}

\noindent \textbf{GGatrieval consistently outperforms baseline methods on the extended dataset.} To thoroughly evaluate the effectiveness of GGatrieval, compared to LLatrieval \citep{li-etal-2024-llatrieval}, we extend our assessment to its retrieval capabilities using the NQ dataset. The results, presented in Table 2, show that retrieval-augmented systems perform consistently well on the single-hop NQ dataset, which involves relatively low reasoning complexity. GGatrieval outperforms baseline models across all evaluation metrics, further validating its robustness and effectiveness across diverse tasks.

\subsection{Ablation Study}

\begin{table}
    \centering
    \resizebox{\linewidth}{!}{ 
    \begin{tabular}{@{}lcccc@{}}
        \hline
        & \multicolumn{2}{c}{ASQA} & \multicolumn{2}{c}{QAMPARI} \\ 
        \cline{2-3} \cline{4-5}
        & EM-R & Cite & F1 & Cite  \\ 
        \hline
        Final Result & 58.30 & 61.30 & 17.90 & 35.42 \\ 
        \textcolor{green}{\textbf{---}} Full Alignment & \textbf{46.79} & \textbf{53.43} & \textbf{13.43} & \textbf{28.52} \\ 
        \textcolor{green}{\textbf{---}} Partial Alignment & 48.38 & 53.13 & 15.04 & 33.65 \\ 
        \textcolor{green}{\textbf{---}} No Alignment & 50.14 & 56.14 & 16.60 & 34.88 \\ 
        \hline
    \end{tabular}
    }
    \caption{Ablation study for excluding documents with different labels. "\textcolor{green}{\textbf{---}}" signifies that a specific category of document has been eliminated. The bolded numbers indicate the worst performance.}
    \label{tab3}
\end{table}

\noindent \textbf{Impact of each alignment label:} This study evaluates the effectiveness of the proposed alignment labels through ablation experiments. As shown in Table 3, we systematically remove documents labeled as Full Alignment, Partial Alignment, and No Alignment from the final selection and assessed model performance on the ASQA and QAMPARI datasets. To ensure a fair comparison, the remaining documents are re-ranked based on alignment and relevance, with the highest-ranked documents filling any gaps. The results reveal that removing more aligned documents leads to a significant decline in Correctness and Citation F1 scores, underscoring the critical role of the FGA strategy. Notably, removing Full Alignment documents have the most pronounced impact on system performance, emphasizing the importance of our selection criterion. 

\noindent \textbf{Interactions between components during iterations:} To explore the relationships between query updates, document retrieval, and system performance, we conduct four rounds of iterative testing using the ASQA and QAMPARI datasets. Figure 3 illustrates the evolution of system performance, the number of updated queries, and the distribution of Full Alignment documents across iterations. The most significant improvement in system performance occurs in the second iteration, which coincides with the greatest decrease in samples containing zero Full Alignment documents and the largest increase in those containing five Full Alignment documents. As the iterations progress, system performance gradually stabilizes. Notably, the performance improvements are closely correlated with the increasing density of Full Alignment documents, underscoring the critical role of the proposed alignment standard in enhancing system performance. Interestingly, no clear pattern emerges in the variation of updated queries, suggesting that the SCQU strategy primarily enhances performance by facilitating the retrieval of a higher number of highly aligned documents. Moreover, within each dataset's set of 1,000 samples, fewer than 30 contained enough Full Alignment documents, while over 600 have none. Based on these observations, we hypothesize that further performance improvements could be achieved by retrieving more Full Alignment documents.    

\begin{figure} 
  \centering 
  \includegraphics[width=\columnwidth]{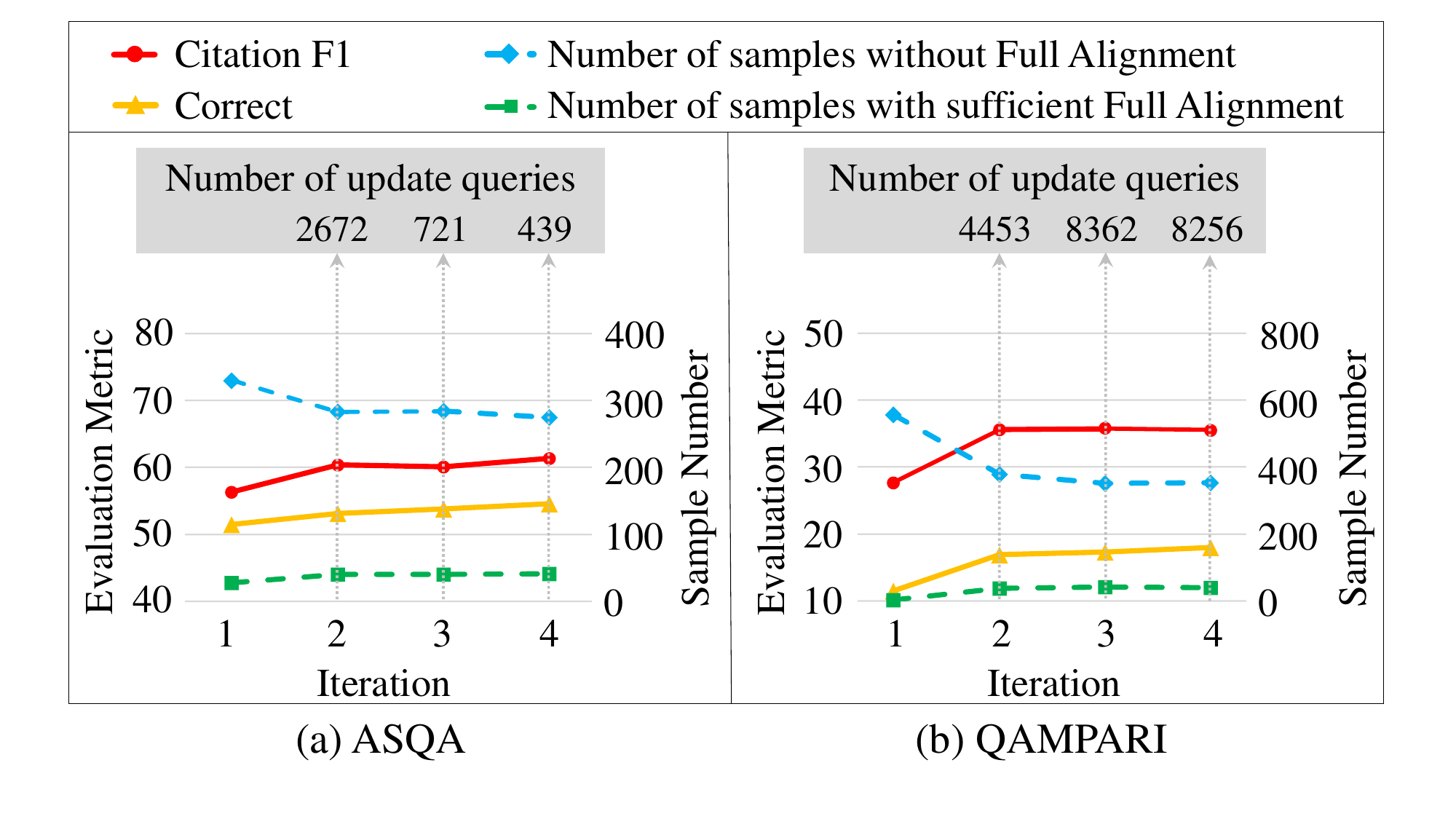} 
  \caption{Impact of interactions among components. The solid line represents the evaluation metric, while the dashed line indicates the distribution of Full Alignment labels within the sample.}
  \label{fig:paper1} 
\end{figure}

\begin{table}[H]
    \centering
    \resizebox{\linewidth}{!}{ 
    \begin{tabular}{@{}lccc@{}} 
        \hline
        & \multicolumn{1}{c}{ASQA}  
        & \multicolumn{1}{c}{QAMPARI}  
        & \multicolumn{1}{c}{ELI5} \\
        \hline
        NA in all docs & 10100 & 25736 & 58128  \\ 
        NA in final docs & 1712 & 1915 & 1153  \\  
        \hline
        PA in all docs & 6081 & 10715 & 53438  \\ 
        PA in final docs & 1663 & 1594 & 2571  \\  
        \hline
        FA in all docs & 2596 & 3386 & 12418  \\ 
        FA in final docs & 1305 & 1337 & 1239  \\ 
        \hline
        NA conversion rate & 0.17 & 0.07 & 0.02  \\
        PA conversion rate & 0.27 & 0.15 & 0.05  \\
        FA conversion rate & \textbf{0.5} & \textbf{0.38} & \textbf{0.1}  \\
        \hline
    \end{tabular}
    }
    \caption{Number and conversion rate of different alignment labels. where "NA" represents "No Alignment", PA represents "Partial Alignment", and FA represents "Full Alignment".}
    \label{tab4}
\end{table}

\subsection{Analysis of Proportions for Different Alignment Labels}

We analyze the proportion of alignment labels in the final selection of documents, as shown in Table 4. In the ASQA, QAMPARI, and ELI5 datasets, Full Alignment and Partial Alignment documents have not dominated the final selection. This observation could be primarily due to variations in label distributions across samples and the relatively low number of Full Alignment documents. However, Full Alignment documents consistently exhibit the highest conversion rates across all three datasets. Additionally, as shown in Figure 4, the proportion of Partial Alignment documents has been notably higher in the ELI5 dataset compared to the ASQA and QAMPARI datasets, while the proportion of No Alignment documents has highly decreased. These results further emphasize the critical role of Full Alignment labels. Moreover, Partial Alignment labels are found to improve the system's robustness, while No Alignment labels effectively serve as a key criterion for document exclusion.

\begin{figure} 
  \centering 
  \includegraphics[width=\columnwidth]{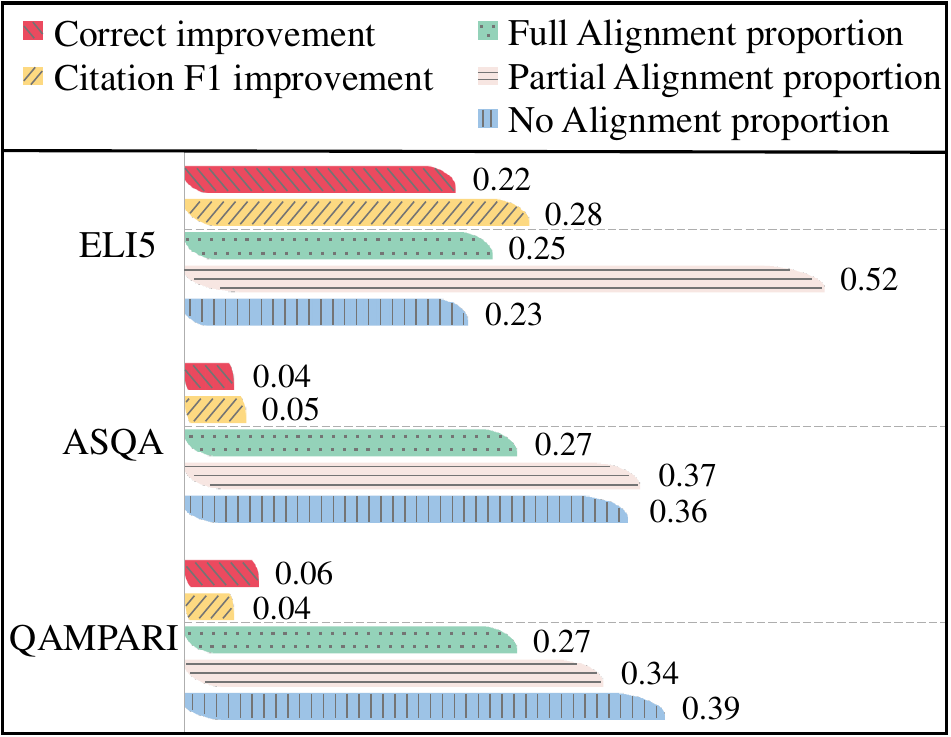} 
  \caption{Cross-dataset analysis of label proportions. The bars above the dashed line represent system performance, while the bars below the dashed line indicate the proportion of documents with different labels.}
  \label{fig:paper1} 
\end{figure}

\section{Conclusion}
We introduce GGatrieval to enhance the retrieval process within the context of verifiable generation.  Unlike traditional methods, our approach simulates the human cognitive process of document retrieval and proposes novel criteria for selecting retrieved documents. The criterion is then used to categorize the documents, and based on this categorization, we present two strategies: the SCQU and the FGA. Together, these strategies optimize the retrieval mechanism, ensuring retrieved documents satisfy verifiability standards and improve overall system performance. Experimental results show that GGatrieval outperforms existing methods, achieving state-of-the-art results.   

\section*{Limitations}

Our approach semantically aligns retrieved documents with the syntactic components of the query but does not model the logical relationships between these components. Moreover, leveraging large language models for semantic alignment introduces significant computational and time costs. Future work could explore applying reinforcement learning or deep learning techniques to address these challenges. Additionally, further investigation is required to understand why a few Full Alignment documents do not contribute to the final set of supporting documents and the factors that underpin the robustness of Partial Alignment documents.

\bibliography{ggatrieval_acl}

\begin{thebibliography}{47}
\providecommand{\natexlab}[1]{#1}

\bibitem[{Amouyal et~al.(2023)Amouyal, Wolfson, Rubin, Yoran, Herzig, and Berant}]{amouyal-etal-2023-qampari}
Samuel Amouyal, Tomer Wolfson, Ohad Rubin, Ori Yoran, Jonathan Herzig, and Jonathan Berant. 2023.
\newblock \href {2023.gem-1.9/} {{QAMPARI}: A benchmark for open-domain questions with many answers}.
\newblock pages 97--110, Singapore.

\bibitem[{Anand et~al.(2023)Anand, Venktesh, Setty, and Anand}]{ref41}
Abhijit Anand, V~Venktesh, Vinay Setty, and Avishek Anand. 2023.
\newblock Context aware query rewriting for text rankers using llm.
\newblock \emph{CoRR}.

\bibitem[{Arora et~al.(2023)Arora, Kini, Chowdhury, Natarajan, Sinha, and Sharma}]{ref77}
Daman Arora, Anush Kini, Sayak~Ray Chowdhury, Nagarajan Natarajan, Gaurav Sinha, and Amit Sharma. 2023.
\newblock Gar-meets-rag paradigm for zero-shot information retrieval.
\newblock \emph{arXiv preprint arXiv:2310.20158}.

\bibitem[{Cheng et~al.(2024)Cheng, Luo, Chen, Liu, Zhao, and Yan}]{ref71}
Xin Cheng, Di~Luo, Xiuying Chen, Lemao Liu, Dongyan Zhao, and Rui Yan. 2024.
\newblock Lift yourself up: Retrieval-augmented text generation with self-memory.
\newblock \emph{Advances in Neural Information Processing Systems}, 36.

\bibitem[{Drozdov et~al.(2022)Drozdov, Sch{\"a}rli, Aky{\"u}rek, Scales, Song, Chen, Bousquet, and Zhou}]{ref1}
Andrew Drozdov, Nathanael Sch{\"a}rli, Ekin Aky{\"u}rek, Nathan Scales, Xinying Song, Xinyun Chen, Olivier Bousquet, and Denny Zhou. 2022.
\newblock Compositional semantic parsing with large language models.
\newblock In \emph{The Eleventh International Conference on Learning Representations}.

\bibitem[{Fan et~al.(2019)Fan, Jernite, Perez, Grangier, Weston, and Auli}]{fan-etal-2019-eli5}
Angela Fan, Yacine Jernite, Ethan Perez, David Grangier, Jason Weston, and Michael Auli. 2019.
\newblock \href {https://doi.org/10.18653/v1/P19-1346} {{ELI}5: Long form question answering}.
\newblock pages 3558--3567, Florence, Italy.

\bibitem[{Gao et~al.()Gao, Chang, Cardie, Brantley, and Joachims}]{ref43}
Ge~Gao, Jonathan~Daniel Chang, Claire Cardie, Kiant{\'e} Brantley, and Thorsten Joachims.
\newblock Policy-gradient training of language models for ranking.
\newblock In \emph{NeurIPS 2023 Foundation Models for Decision Making Workshop}.

\bibitem[{Gao et~al.(2023)Gao, Yen, Yu, and Chen}]{gao-etal-2023-enabling}
Tianyu Gao, Howard Yen, Jiatong Yu, and Danqi Chen. 2023.
\newblock \href {https://doi.org/10.18653/v1/2023.emnlp-main.398} {Enabling large language models to generate text with citations}.
\newblock pages 6465--6488, Singapore.

\bibitem[{Gupta et~al.(2024)Gupta, Ranjan, and Singh}]{ref69}
Shailja Gupta, Rajesh Ranjan, and Surya~Narayan Singh. 2024.
\newblock A comprehensive survey of retrieval-augmented generation (rag): Evolution, current landscape and future directions.
\newblock \emph{arXiv preprint arXiv:2410.12837}.

\bibitem[{Hei et~al.(2024)Hei, Liu, Ou, Qiao, Jiao, Song, Tian, and Lin}]{ref32}
Zijian Hei, Weiling Liu, Wenjie Ou, Juyi Qiao, Junming Jiao, Guowen Song, Ting Tian, and Yi~Lin. 2024.
\newblock Dr-rag: Applying dynamic document relevance to retrieval-augmented generation for question-answering.
\newblock \emph{arXiv preprint arXiv:2406.07348}.

\bibitem[{Izacard et~al.(2023)Izacard, Lewis, Lomeli, Hosseini, Petroni, Schick, Dwivedi-Yu, Joulin, Riedel, and Grave}]{ref70}
Gautier Izacard, Patrick Lewis, Maria Lomeli, Lucas Hosseini, Fabio Petroni, Timo Schick, Jane Dwivedi-Yu, Armand Joulin, Sebastian Riedel, and Edouard Grave. 2023.
\newblock Atlas: Few-shot learning with retrieval augmented language models.
\newblock \emph{Journal of Machine Learning Research}, 24(251):1--43.

\bibitem[{Kaplan et~al.(2020)Kaplan, McCandlish, Henighan, Brown, Chess, Child, Gray, Radford, Wu, and Amodei}]{ref53}
Jared Kaplan, Sam McCandlish, Tom Henighan, Tom~B Brown, Benjamin Chess, Rewon Child, Scott Gray, Alec Radford, Jeffrey Wu, and Dario Amodei. 2020.
\newblock Scaling laws for neural language models.
\newblock \emph{arXiv preprint arXiv:2001.08361}.

\bibitem[{Karpukhin et~al.(2020)Karpukhin, Oguz, Min, Lewis, Wu, Edunov, Chen, and Yih}]{karpukhin-etal-2020-dense}
Vladimir Karpukhin, Barlas Oguz, Sewon Min, Patrick Lewis, Ledell Wu, Sergey Edunov, Danqi Chen, and Wen-tau Yih. 2020.
\newblock \href {https://doi.org/10.18653/v1/2020.emnlp-main.550} {Dense passage retrieval for open-domain question answering}.
\newblock pages 6769--6781, Online.

\bibitem[{Khandelwal et~al.(2019)Khandelwal, Levy, Jurafsky, Zettlemoyer, and Lewis}]{ref60}
Urvashi Khandelwal, Omer Levy, Dan Jurafsky, Luke Zettlemoyer, and Mike Lewis. 2019.
\newblock Generalization through memorization: Nearest neighbor language models.
\newblock In \emph{International Conference on Learning Representations}.

\bibitem[{Kwiatkowski et~al.(2019)Kwiatkowski, Palomaki, Redfield, Collins, Parikh, Alberti, Epstein, Polosukhin, Devlin, Lee, Toutanova, Jones, Kelcey, Chang, Dai, Uszkoreit, Le, and Petrov}]{kwiatkowski-etal-2019-natural}
Tom Kwiatkowski, Jennimaria Palomaki, Olivia Redfield, Michael Collins, Ankur Parikh, Chris Alberti, Danielle Epstein, Illia Polosukhin, Jacob Devlin, Kenton Lee, Kristina Toutanova, Llion Jones, Matthew Kelcey, Ming-Wei Chang, Andrew~M. Dai, Jakob Uszkoreit, Quoc Le, and Slav Petrov. 2019.
\newblock \href {https://doi.org/10.1162/tacl_a_00276} {Natural questions: A benchmark for question answering research}.
\newblock \emph{Transactions of the Association for Computational Linguistics}, 7:452--466.

\bibitem[{Lee et~al.(2023)Lee, Whang, Lee, and Lim}]{ref22}
Dongyub Lee, Taesun Whang, Chanhee Lee, and Heuiseok Lim. 2023.
\newblock Towards reliable and fluent large language models: Incorporating feedback learning loops in qa systems.
\newblock \emph{arXiv preprint arXiv:2309.06384}.

\bibitem[{Lesmo and Lombardo(1992)}]{lesmo-lombardo-1992-assignment}
Leonardo Lesmo and Vincenzo Lombardo. 1992.
\newblock \href {C92-4170/} {The assignment of grammatical relations in natural language processing}.

\bibitem[{Lewis et~al.(2020)Lewis, Perez, Piktus, Petroni, Karpukhin, Goyal, K{\"u}ttler, Lewis, Yih, Rockt{\"a}schel et~al.}]{ref6}
Patrick Lewis, Ethan Perez, Aleksandra Piktus, Fabio Petroni, Vladimir Karpukhin, Naman Goyal, Heinrich K{\"u}ttler, Mike Lewis, Wen-tau Yih, Tim Rockt{\"a}schel, et~al. 2020.
\newblock Retrieval-augmented generation for knowledge-intensive nlp tasks.
\newblock \emph{Advances in Neural Information Processing Systems}, 33:9459--9474.

\bibitem[{Li et~al.(2023)Li, Liu, Xiao, and Shao}]{ref3}
Chaofan Li, Zheng Liu, Shitao Xiao, and Yingxia Shao. 2023.
\newblock Making large language models a better foundation for dense retrieval.
\newblock \emph{arXiv preprint arXiv:2312.15503}.

\bibitem[{Li et~al.(2024{\natexlab{a}})Li, Yuan, and Zhang}]{ref76}
Jiarui Li, Ye~Yuan, and Zehua Zhang. 2024{\natexlab{a}}.
\newblock Enhancing llm factual accuracy with rag to counter hallucinations: A case study on domain-specific queries in private knowledge-bases.
\newblock \emph{arXiv preprint arXiv:2403.10446}.

\bibitem[{Li et~al.(2025)Li, Stenzel, Eickhoff, and Bahrainian}]{ref55}
Siran Li, Linus Stenzel, Carsten Eickhoff, and Seyed~Ali Bahrainian. 2025.
\newblock Enhancing retrieval-augmented generation: A study of best practices.
\newblock In \emph{Proceedings of the 31st International Conference on Computational Linguistics}, pages 6705--6717.

\bibitem[{Li et~al.(2024{\natexlab{b}})Li, Zhu, Li, Yin, Sun, and Qiu}]{li-etal-2024-llatrieval}
Xiaonan Li, Changtai Zhu, Linyang Li, Zhangyue Yin, Tianxiang Sun, and Xipeng Qiu. 2024{\natexlab{b}}.
\newblock \href {https://doi.org/10.18653/v1/2024.naacl-long.305} {{LL}atrieval: {LLM}-verified retrieval for verifiable generation}.
\newblock pages 5453--5471, Mexico City, Mexico.

\bibitem[{Liu et~al.(2024)Liu, AlDahoul, Eady, Zaki, AlShebli, and Rahwan}]{ref4}
Fengyuan Liu, Nouar AlDahoul, Gregory Eady, Yasir Zaki, Bedoor AlShebli, and Talal Rahwan. 2024.
\newblock Self-reflection outcome is sensitive to prompt construction.
\newblock \emph{arXiv preprint arXiv:2406.10400}.

\bibitem[{Liu and Mozafari(2024)}]{ref42}
Jie Liu and Barzan Mozafari. 2024.
\newblock Query rewriting via large language models.
\newblock \emph{arXiv preprint arXiv:2403.09060}.

\bibitem[{Liu et~al.(2023)Liu, Lai, Yu, Xu, Zeng, Du, Zhang, Dong, and Tang}]{ref52}
Xiao Liu, Hanyu Lai, Hao Yu, Yifan Xu, Aohan Zeng, Zhengxiao Du, Peng Zhang, Yuxiao Dong, and Jie Tang. 2023.
\newblock Webglm: Towards an efficient web-enhanced question answering system with human preferences.
\newblock In \emph{Proceedings of the 29th ACM SIGKDD Conference on Knowledge Discovery and Data Mining}, pages 4549--4560.

\bibitem[{Ma et~al.(2023)Ma, Gong, He, Zhao, and Duan}]{ma-etal-2023-query}
Xinbei Ma, Yeyun Gong, Pengcheng He, Hai Zhao, and Nan Duan. 2023.
\newblock \href {https://doi.org/10.18653/v1/2023.emnlp-main.322} {Query rewriting in retrieval-augmented large language models}.
\newblock pages 5303--5315, Singapore.

\bibitem[{Mao et~al.(2024)Mao, Jiang, Chen, Li, Wang, Wang, Xie, Huang, Chen, and Zhang}]{ref33}
Shengyu Mao, Yong Jiang, Boli Chen, Xiao Li, Peng Wang, Xinyu Wang, Pengjun Xie, Fei Huang, Huajun Chen, and Ningyu Zhang. 2024.
\newblock Rafe: Ranking feedback improves query rewriting for rag.
\newblock \emph{arXiv preprint arXiv:2405.14431}.

\bibitem[{Min et~al.(2020)Min, Michael, Hajishirzi, and Zettlemoyer}]{min-etal-2020-ambigqa}
Sewon Min, Julian Michael, Hannaneh Hajishirzi, and Luke Zettlemoyer. 2020.
\newblock \href {https://doi.org/10.18653/v1/2020.emnlp-main.466} {{A}mbig{QA}: Answering ambiguous open-domain questions}.
\newblock pages 5783--5797, Online.

\bibitem[{Nakano et~al.(2021)Nakano, Hilton, Balaji, Wu, Ouyang, Kim, Hesse, Jain, Kosaraju, Saunders et~al.}]{ref26}
Reiichiro Nakano, Jacob Hilton, Suchir Balaji, Jeff Wu, Long Ouyang, Christina Kim, Christopher Hesse, Shantanu Jain, Vineet Kosaraju, William Saunders, et~al. 2021.
\newblock Webgpt: Browser-assisted question-answering with human feedback.
\newblock \emph{arXiv preprint arXiv:2112.09332}.

\bibitem[{Piktus et~al.(2021)Piktus, Petroni, Karpukhin, Okhonko, Broscheit, Izacard, Lewis, O{\u{g}}uz, Grave, Yih et~al.}]{ref16}
Aleksandra Piktus, Fabio Petroni, Vladimir Karpukhin, Dmytro Okhonko, Samuel Broscheit, Gautier Izacard, Patrick Lewis, Barlas O{\u{g}}uz, Edouard Grave, Wen-tau Yih, et~al. 2021.
\newblock The web is your oyster-knowledge-intensive nlp against a very large web corpus.
\newblock \emph{arXiv preprint arXiv:2112.09924}.

\bibitem[{Raina and Gales(2024)}]{raina-gales-2024-question}
Vatsal Raina and Mark Gales. 2024.
\newblock \href {https://doi.org/10.18653/v1/2024.fever-1.25} {Question-based retrieval using atomic units for enterprise {RAG}}.
\newblock pages 219--233, Miami, Florida, USA.

\bibitem[{Robertson et~al.(2009)Robertson, Zaragoza et~al.}]{ref17}
Stephen Robertson, Hugo Zaragoza, et~al. 2009.
\newblock The probabilistic relevance framework: Bm25 and beyond.
\newblock \emph{Foundations and Trends{\textregistered} in Information Retrieval}, 3(4):333--389.

\bibitem[{Saad-Falcon et~al.(2023)Saad-Falcon, Khattab, Santhanam, Florian, Franz, Roukos, Sil, Sultan, and Potts}]{saad-falcon-etal-2023-udapdr}
Jon Saad-Falcon, Omar Khattab, Keshav Santhanam, Radu Florian, Martin Franz, Salim Roukos, Avirup Sil, Md~Sultan, and Christopher Potts. 2023.
\newblock \href {https://doi.org/10.18653/v1/2023.emnlp-main.693} {{UDAPDR}: Unsupervised domain adaptation via {LLM} prompting and distillation of rerankers}.
\newblock pages 11265--11279, Singapore.

\bibitem[{Sarthi et~al.()Sarthi, Abdullah, Tuli, Khanna, Goldie, and Manning}]{ref73}
Parth Sarthi, Salman Abdullah, Aditi Tuli, Shubh Khanna, Anna Goldie, and Christopher~D Manning.
\newblock Raptor: Recursive abstractive processing for tree-organized retrieval.
\newblock In \emph{The Twelfth International Conference on Learning Representations}.

\bibitem[{Stelmakh et~al.(2022)Stelmakh, Luan, Dhingra, and Chang}]{stelmakh-etal-2022-asqa}
Ivan Stelmakh, Yi~Luan, Bhuwan Dhingra, and Ming-Wei Chang. 2022.
\newblock \href {https://doi.org/10.18653/v1/2022.emnlp-main.566} {{ASQA}: Factoid questions meet long-form answers}.
\newblock pages 8273--8288, Abu Dhabi, United Arab Emirates.

\bibitem[{Sun et~al.(2023)Sun, Yan, Ma, Wang, Ren, Chen, Yin, and Ren}]{sun-etal-2023-chatgpt}
Weiwei Sun, Lingyong Yan, Xinyu Ma, Shuaiqiang Wang, Pengjie Ren, Zhumin Chen, Dawei Yin, and Zhaochun Ren. 2023.
\newblock \href {https://doi.org/10.18653/v1/2023.emnlp-main.923} {Is {C}hat{GPT} good at search? investigating large language models as re-ranking agents}.
\newblock pages 14918--14937, Singapore.

\bibitem[{Tang et~al.(2024)Tang, Chen, Yu, Lu, Fu, Yu, Lin, Huang, He, Han et~al.}]{ref34}
Qiaoyu Tang, Jiawei Chen, Bowen Yu, Yaojie Lu, Cheng Fu, Haiyang Yu, Hongyu Lin, Fei Huang, Ben He, Xianpei Han, et~al. 2024.
\newblock Self-retrieval: Building an information retrieval system with one large language model.
\newblock \emph{arXiv preprint arXiv:2403.00801}.

\bibitem[{Thoppilan et~al.(2022)Thoppilan, De~Freitas, Hall, Shazeer, Kulshreshtha, Cheng, Jin, Bos, Baker, Du et~al.}]{ref28}
Romal Thoppilan, Daniel De~Freitas, Jamie Hall, Noam Shazeer, Apoorv Kulshreshtha, Heng-Tze Cheng, Alicia Jin, Taylor Bos, Leslie Baker, Yu~Du, et~al. 2022.
\newblock Lamda: Language models for dialog applications.
\newblock \emph{arXiv preprint arXiv:2201.08239}.

\bibitem[{Tian et~al.(2024)Tian, Xia, and Song}]{tian-etal-2024-large}
Yuanhe Tian, Fei Xia, and Yan Song. 2024.
\newblock \href {https://doi.org/10.18653/v1/2024.acl-long.384} {Large language models are no longer shallow parsers}.
\newblock pages 7131--7142, Bangkok, Thailand.

\bibitem[{Wang et~al.(2024)Wang, Teo, Ouyang, Xu, and Shi}]{wang-etal-2024-rag}
Zheng Wang, Shu Teo, Jieer Ouyang, Yongjun Xu, and Wei Shi. 2024.
\newblock \href {https://doi.org/10.18653/v1/2024.acl-long.108} {{M}-{RAG}: Reinforcing large language model performance through retrieval-augmented generation with multiple partitions}.
\newblock pages 1966--1978, Bangkok, Thailand.

\bibitem[{Wang et~al.(2023)Wang, Araki, Jiang, Parvez, and Neubig}]{ref66}
Zhiruo Wang, Jun Araki, Zhengbao Jiang, Md~Rizwan Parvez, and Graham Neubig. 2023.
\newblock Learning to filter context for retrieval-augmented generation.
\newblock \emph{arXiv preprint arXiv:2311.08377}.

\bibitem[{Weller et~al.(2024)Weller, Marone, Weir, Lawrie, Khashabi, and Van~Durme}]{weller-etal-2024-according}
Orion Weller, Marc Marone, Nathaniel Weir, Dawn Lawrie, Daniel Khashabi, and Benjamin Van~Durme. 2024.
\newblock \href {2024.eacl-long.140/} {{\textquotedblleft}according to . . . {\textquotedblright}: Prompting language models improves quoting from pre-training data}.
\newblock pages 2288--2301, St. Julian{'}s, Malta.

\bibitem[{Xiao et~al.(2024)Xiao, Liu, Zhang, Muennighoff, Lian, and Nie}]{ref15}
Shitao Xiao, Zheng Liu, Peitian Zhang, Niklas Muennighoff, Defu Lian, and Jian-Yun Nie. 2024.
\newblock C-pack: Packed resources for general chinese embeddings.
\newblock In \emph{Proceedings of the 47th international ACM SIGIR conference on research and development in information retrieval}, pages 641--649.

\bibitem[{Yan et~al.(2024)Yan, Gu, Zhu, and Ling}]{ref31}
Shi-Qi Yan, Jia-Chen Gu, Yun Zhu, and Zhen-Hua Ling. 2024.
\newblock Corrective retrieval augmented generation.
\newblock \emph{arXiv preprint arXiv:2401.15884}.

\bibitem[{Yao et~al.(2023)Yao, Zhao, Yu, Du, Shafran, Narasimhan, and Cao}]{ref72}
Shunyu Yao, Jeffrey Zhao, Dian Yu, Nan Du, Izhak Shafran, Karthik Narasimhan, and Yuan Cao. 2023.
\newblock React: Synergizing reasoning and acting in language models.
\newblock In \emph{International Conference on Learning Representations (ICLR)}.

\bibitem[{Zhang et~al.(2024)Zhang, Yu, Wang, and Zhang}]{zhang-etal-2024-arl2}
LingXi Zhang, Yue Yu, Kuan Wang, and Chao Zhang. 2024.
\newblock \href {https://doi.org/10.18653/v1/2024.acl-long.203} {{ARL}2: Aligning retrievers with black-box large language models via self-guided adaptive relevance labeling}.
\newblock pages 3708--3719, Bangkok, Thailand.

\bibitem[{Zhou et~al.(2024)Zhou, Meng, Jin, and Han}]{zhou-etal-2024-grasping}
Sizhe Zhou, Yu~Meng, Bowen Jin, and Jiawei Han. 2024.
\newblock \href {https://doi.org/10.18653/v1/2024.emnlp-main.747} {Grasping the essentials: Tailoring large language models for zero-shot relation extraction}.
\newblock pages 13462--13486, Miami, Florida, USA.

\end{thebibliography}

\end{document}